\documentclass{article}
\usepackage[preprint,nonatbib]{neurips_2024}
\usepackage[utf8]{inputenc} 
\usepackage[T1]{fontenc}    
\usepackage{hyperref}       
\usepackage{url}            
\usepackage{booktabs}       
\usepackage{amsfonts}       
\usepackage{nicefrac}       
\usepackage{microtype}      
\usepackage{xcolor}         
\usepackage{mathtools}

\title{Synthetic Data Privacy Metrics}

\author{
Amy Steier \\
Gretel.ai \\
 \texttt{amy@gretel.ai} \\
\And
Lipika Ramaswamy \\
Gretel.ai \\
 \texttt{lipika@gretel.ai} \\
 \And
 Andre Manoel \\
 Gretel.ai \\
  \texttt{andre.manoel@gretel.ai} \\
\And
Alexa Haushalter \\
Gretel.ai \\
\texttt{alexa@gretel.ai} \\
}

\begin{document}

\maketitle

\begin{abstract}
Recent advancements in generative AI have made it possible to create synthetic datasets that can be as accurate as real-world data for training AI models, powering statistical insights, and fostering collaboration with sensitive datasets while offering strong privacy guarantees. Effectively measuring the empirical privacy of synthetic data is an important step in the process. However, while there is a multitude of new privacy metrics being published every day, there currently is no standardization. In this paper, we review  the pros and cons of popular metrics that include simulations of adversarial attacks.  We also review current best practices for amending generative models to enhance the privacy of the data they create (e.g. differential privacy). 
\end{abstract}

\section{Introduction}
Fundamentally, synthetic data is annotated information that computer simulations or algorithms generate as an alternative to real-world data \cite{1}. It is generated using a class of machine learning models that learn the underlying relationships in a real-world dataset and can generate new data instances. High-quality synthetic data will contain the same statistical relationship between variables (e.g., if real-world data reveals that weight and height are positively correlated, then this observation will hold true in the synthetic data). However, records of a synthetic dataset will not have a clear link to records of the original, real-world data on which it is based. Synthetic data should be one step removed from real-world data. This means that synthetic data generated with the right level of privacy protection can provide safeguards against adversarial attacks, something traditional anonymization techniques like masking or tokenization cannot promise.

Synthetic data has been hailed as a solution to the difficult problem of sharing sensitive data while protecting privacy. However, the fact that synthetic records are “artificial” does not, per se, guarantee that the privacy of all individuals in the original dataset is protected. There is abundant evidence that synthetic generative models can suffer from unintended memorization \cite{2} and can leak information associated with training samples \cite{3, 4, 5, 71, 74}. Thus effective privacy metrics quantify such leakage and are a critical part of the use of synthetic data.

There is currently a lack of standardization of  empirical privacy metrics for synthetic data. In this paper, we review popular metrics as well as techniques for strengthening the privacy of synthetic data.

\section{Background}
\subsection{Data types Under Consideration}
The dominant type of synthetic data is tabular, but it is also common to see synthetic text, image, graph, audio and video data.  This paper focuses on tabular synthetic data. The privacy metrics that exploit adversarial attacks have also been used with synthetic text and image data. These are typically membership inference attacks (MIAs) and attribute inference attacks (AIAs). With text data it is also common to see measurements of how often personally identifiable information (PII) is leaked from the original to the synthetic data.

\subsection{The Privacy-Utility Trade-off}
The privacy-utility trade-off in synthetic data refers to the balance between ensuring the privacy of individuals' information and maintaining the usefulness of the generated data for analysis or training a downstream model. Techniques to enhance the privacy of synthetic data can sometimes come at the cost of decreasing utility. The measurement of  the quality and utility of synthetic data has been well studied \cite{59, 60, 61}. This paper focuses specifically on privacy metrics.

\section{Privacy Metrics}
In this section we review the popular privacy metrics of k-anonymity, personally identifiable information (PII)  replay, exact match counts, distance to closest record (DCR), nearest neighbor distance ratio (NNDR), nearest neighbor adversarial accuracy (NNAA), membership inference attacks (MIAs) and attribute inference attacks (AIAs).

\subsection{K-Anonymity}
K-anonymity is a popular privacy metric that dates back to 1986 \cite{6} but the term was coined in a 1998 paper \cite{7}. Data is said to have the k-anonymity property if the information for each person contained in the data cannot be distinguished from at least k - 1 individuals whose information also appears in the data.  K-anonymity protects against hackers or malicious parties using ‘re-identification,’ or the practice of tracing data’s origins back to the individual it is connected to in the real world. Variants of k-anonymity include l-diversity and t-closeness \cite{108}.

To assess k-anonymity one must first determine whether each field is a direct identifier, non-identifier or  quasi-identifiers. Examples of direct identifiers are name or address. In k-anonymity, it is presumed that the direct identifiers will be suppressed and it's the quasi-identifiers that are used by an adversary for re-identification. Quasi-identifiers are semi-identifying fields that somehow an adversary knows. For example, a date of birth is often a quasi-identifier because it is information about an individual that is known or relatively easy for an adversary to find out. More generally, an adversary may know the quasi-identifiers about an individual because that individual is an acquaintance of the adversary or because the adversary has access to a population database or registry of identifiable information. Research shows that the combination of quasi-identifiers ‘gender’, ‘birth date’, and ‘postal code’ reidentify between 63 and 87\% of the U.S. population \cite{72}.

Although k-anonymity is a popular and easy to understand metric, pragmatically it can be difficult to know which fields are quasi-identifiers.  By definition, a quasi-identifier is a field that when linked to external information could be used to identify an individual. To some extent, almost any field could be a quasi-identifier making it difficult for practitioners to know which fields to use. This measure also tends to not scale well with really high dimensional data \cite{8}.

\subsection{Personally Identifiable Information (PII) Replay}
Personally identifiable information (PII) in natural language is data that can re-identify an individual. PII can be a direct identifier or quasi-identifier. The hope in creating synthetic data is that PII in text won't be replayed in the synthetic data. Measuring PII replay as a metric is important because without added precautions, it can be a common occurrence \cite{9, 10, 11}. It should be noted, however, that different types of PII carry different risks.  If a common first name or last name is replayed there's little privacy risk, but if a full name is replayed the risk is higher.

Typical approaches for reducing PII replay are Differential Privacy \cite{12}, PII scrubbing \cite{9} or pseudonymization \cite{13}. Read more about such recourse measures in Section 4. 

\subsection{Exact Match Counts}
This metric, sometimes referred to as Identical Match Share (IMS), measures the number of training records that are exactly replicated in the synthetic data. Theoretically, you're shooting for no exact matches. In the ideal privacy/utility tradeoff scenario, the goal is to have the generated data be in the "Goldilocks" zone: not too similar to the training data, but also not too dissimilar. 

IMS captures the proportion of identical copies between train (original) and synthetic records. The test passes if that proportion is smaller or equal to the one between train and test datasets. A test dataset is the portion of the original data put aside before a synthetic model is trained. Assessing exact matches is a common practice in industry \cite{14,15,16,17}.

\subsection{Distance to Closest Record (DCR)}
Distance to closest record (DCR) is a popular privacy metric that has been in use for quite some time \cite{18}. There are many variations of it. In its most basic form it measures the Euclidean distance between any synthetic record and its closest corresponding real neighbor.  Sometimes the median distance is reported \cite{19} and sometimes distance at the 5th percentile is used \cite{20}. Ideally, the higher the DCR the lesser the risk of privacy breach. DCR of 0 means it replicated training examples in the synthetic data. DCR is supposed to protect against settings where the train data was just slightly perturbed or noised and presented as synthetic.

One variation of DCR is to compare the train-train DCR with the train-synth DCR \cite{15,19}. Privacy is preserved if train-train DCR is less than train-synth DCR. This means the synthetic data is further from the real data than the real data is to itself, meaning that the synthetic data is more private than a copy or a simple perturbation of the real data.

Another variation is to look at DCR within real data and DCR within synthetic data.  The former metric is the Euclidean distance between each record in the real data and its closest corresponding record in the real data, whereas the latter measures the same but on the synthetic dataset. If the DCR within the synthetic data is less than that in the real data, this can be indicative of a model collapse problem \cite{21}.

Finally, yet another variation is to compute the train-synth DCR and the holdout-synth DCR. The holdout dataset is from the same source as the training, but it was not used in the training. If the synthetic data is significantly closer to the training data than the holdout data, this means that some information specific to the training data has leaked into the synthetic dataset. If the synthetic data is significantly farther from the training data than the holdout data, this means that we have lost information in terms of accuracy or fidelity. Sometimes the share of synthetic records that are closer to a training than to a holdout record is computed \cite{24}. A share of  50\% or less indicates that the synthetic dataset does not provide any information that could lead an attacker to assume whether a certain individual was actually present in the training set.

DCR is a quantitative and easily interpretable privacy metric.  It is widely used both in academia \cite{20, 22, 23, 27, 28, 29, 30, 31} and industry \cite{14, 15, 17, 24, 25, 16}.

\subsection{Nearest Neighbor Distance Ratio (NNDR)}
Nearest neighbor distance ratio (NNDR) measures the ratio of each synthetic record's distance to the nearest training neighbor compared to the distance to the second nearest training neighbor. This ratio is within $[0, 1]$. Higher values indicate better privacy. Low values may reveal sensitive information from the closest real data record. A value close to 0 implies that the point is likely to be close to an outlier. Thus, NNDR serves as a measure of proximity to outliers in the original data set and outliers are known to be more vulnerable to adversarial attacks \cite{36, 37, 38, 36}.

There are many similarities to DCR. Both median distance \cite{35} and 5th percentile \cite{20} are commonly used. Further, sometimes the train-synth NNDR is compared to the holdout-synth NNDR \cite{35}. The holdout dataset is from the same source as the training, but it was not used in the training. If the train-synth NNDR is significantly higher than the holdout-synth NNDR this is an indication that information specific to the training set was leaked into the synthetic set. If the train-synth NNDR is significantly lower than the holdout-synth NNDR, this suggests that information has been lost and the synthetic data may not have high fidelity. Further, sometimes the train-train NNDR is compared to the synth-synth NNDR to look for signs of model collapse \cite{34}.

NNDR is a quantitative and easy to understand metric that is common in both industry \cite{14, 32} and academia \cite{20, 33, 34, 27}.

\subsection{Nearest Neighbor Adversarial Accuracy (NNAA)}

Nearest neighbor adversarial accuracy (NNAA) is a privacy metric that directly measures the extent to which a generative model overfits the real dataset \cite{30,41}. This type of measurement is important as adversarial attacks often exploit model overfitting \cite{43, 55}. The equation for adversarial accuracy is shown below.

\[
AA_{TS} = \frac{1}{2}\left(\frac{1}{n}\sum_{i=1}^{n}1(d_{TS}(i) > d_{TT}(i)) + \frac{1}{n}\sum_{i=1}^{n}1(d_{ST}(i) > d_{SS}(i))\right)
\]
In this equation, the distance from one point in a target distribution R is compared to the nearest point in a source distribution S as well as the distance to the next nearest point in the target distribution. These distances are defined as $d_{TS}(i) = min_{j}\parallel{x_{T}^i - x_{S}^{j}}\parallel$ and $d_{TT}(i) = min_{j,j\not=i}\parallel{x_{T}^i - x_{T}^{j}}\parallel$

Privacy loss is then defined by the difference between the adversarial accuracy on the test set and the adversarial accuracy on the training set as shown in the below equations:

\[
(Train Adversarial Acc.) = E[AA_{RtrA_1}]
\]
\[
(Test Adversarial Acc.) = E[AA_{RteA_2}]
\]
\[
\textbf{PrivacyLoss} = Test AA - Train AA
\]

NNAA essentially measures the effectiveness of an adversary in distinguishing between in-distribution and out-of-distribution data points based on their nearest neighbors. Yale et al \cite{41} explain in their paper that if the generative model does a good job, then the adversarial classifier can't distinguish between generated data and real data; train and test adversarial accuracy should both be 0.5, and the privacy loss will be 0. If the generator overfits the training data, the train adversarial accuracy will be near 0 (good training resemblance), but the test adversarial accuracy will be around 0.5 (poor test resemblance). Thus the privacy loss will be high (near 0.5). If the generative model does a poor job and underfits, it will serve generated data that does not resemble real data. Thus the adversarial classifier will have no problem classifying real vs. synthetic so the train and test adversarial accuracy will both be high (>0.5) and similar, and the privacy loss will also be near 0. In this last case, privacy is preserved but the utility of the data may be low.

NNAA is an increasingly referenced privacy metric \cite{42, 60} that can give a user confidence that their data is safe from adversarial attacks.

\subsection{Membership Inference Attacks (MIAs)}
Membership inference attacks (MIAs) have become an increasingly common privacy metric. In a synthetic data context, an MIA is when an attacker tries to identify if certain real records have been used to train the synthetic data generation algorithm. This is a privacy risk since, for example, if the synthetic dataset is about breast cancer, then the attacker can deduce if the person they found has breast cancer.

The first MIA against machine learning models was proposed by Shokri et al \cite{43}. In this attack, the adversary trains shadow models on datasets sampled from the same data distribution as the target model. The adversary then calculates the difference between the perplexity of the target record w.r.t. the target and shadow models, and uses this as a score to decide if the record was a training member or not. The main intuition behind this design is that models behave differently when they see data that do not belong to the training dataset. This difference is captured in the model outputs as well as in their internal representations. The use of shadow models remains the most common form of black box MIAs \cite{44, 46, 47, 48, 49, 50, 51, 52, 53, 54, 58}. LiRA (Likelihood Ratio Attack) by Carlini et al \cite{58} is a popular shadow-based MIA. Shadow models are sometimes referred to as reference models. The number of shadow models affects the attack accuracy, but it also incurs cost to the attackers. Salem et al \cite{44} showed that MIAs are possible with as little as one shadow model. 

In order to train shadow models, attacks of this kind make the strong and arguably unrealistic assumption that an adversary has access to samples closely resembling the original training data. Mattern et al \cite{45} propose and evaluate neighborhood attacks, which compare model scores for a given sample to scores of synthetically generated neighbor texts and therefore eliminate the need for access to the training data distribution.

Another common and highly simple attack is the LOSS attack \cite{55, 56, 57}. These attacks classify samples as training members if their loss values are below a certain threshold. They exploit models’ tendency to overfit their training data and, therefore, exhibit lower loss values for training members.  While these attacks avoid the need to access the training data distribution, they often don't perform as well as attacks that use shadow models.

The above attacks are typically either white-box (adversary has access to model details) or black-box (no model details, but adversaries can query the model and access record losses). However, many industrial settings are essentially no-box. The adversary has access to the synthetic data and that's it. In this environment, distance-based MIAs are often used \cite{59, 60, 61, 62, 63, 64, 65}.  The attacker predicts membership based on the distance between the target record and its nearest neighbor in the synthetic dataset, with an empirically selected threshold. 

\begin{figure}[h]
\includegraphics[scale=.65]{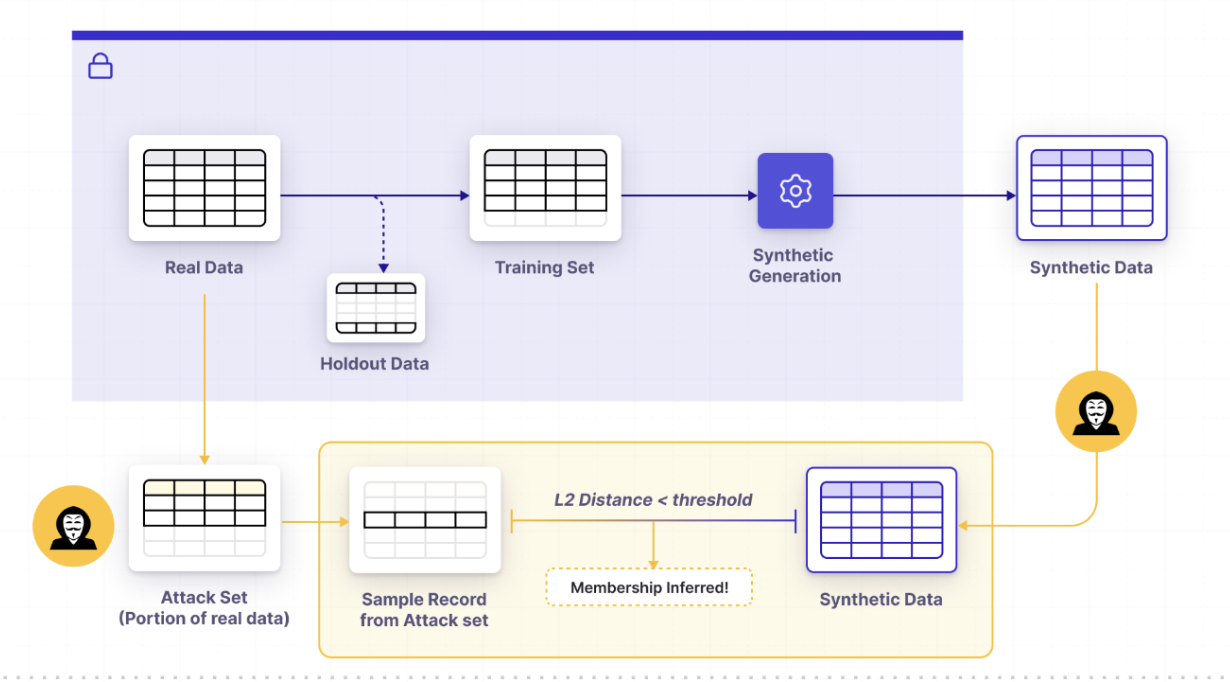}
\caption{No-box Distance Based MIAs}
\end{figure}

Figure 1 demonstrates how a no-box distance-based MIA typically works. To start, 5\% of the training data is put aside in a holdout set and is not used in the fine-tuning. To conduct an MIA simulation, the holdout set is combined with an equivalent-sized sample from the training (fine-tuning) set to create the attack dataset. L2 distance is often used to compare each attack record to the synthetic data and find the nearest neighbor distance. This distance is compared to a threshold value and if it’s less than the threshold it’s called a match. Matches from the training set are considered true positives (TPs) and matches from the holdout set are considered false positives (FPs). 

From here, precision and accuracy can be computed for each attack. Generally speaking, both precision and accuracy should be below 0.5 to get a grade of Excellent. When both precision and accuracy are 0.8 or above that results in a grade of Poor. Scores that fall between 0.5 and 0.8 translate into Moderate, Good or Very Good, depending on where they fall in the range. Many attacks are often simulated each time varying the training sample, distance threshold and proportion of training samples used. An overall score/grade is computed by taking the average of all MIA simulations. 

Distance based MIAs provide a direct measurement of how safe synthetic data is from adversarial attacks. This directness is an advantage over measures like DCR and NNDR which are essentially trying to measure the same thing but in an indirect way.

MIAs are also commonly used with Large Language Models (LLMs) to estimate privacy leakage. Several researchers have found that the pre-training data for LLMs is susceptible to MIAs \cite{66,67}. However, fine-tuning data, which tends to be smaller and in training for more epochs, has been shown to be more vulnerable to MIAs than pre-training data \cite{68, 70}. Finally, in-context learned (ICL) LLMs are more vulnerable to MIAs than fine-tuned LLMs \cite{69}.

MIAs provide a tangible, practical way to measure the privacy risk in synthetic data. Successful MIAs have been shown to be a gateway to additional attacks \cite{71}.

\subsection{Attribute Inference Attacks (AIAs)}
In an attribute inference attack (AIA) the adversary tries to use knowledge of some of the fields (the quasi-identifiers) to determine values for other sensitive fields. Quasi-identifiers are the semi-identifying fields that somehow an adversary has access.  See the section on k-anonymity above for more information on quasi-identifiers. 

\begin{figure}[h]
\includegraphics[scale=.65]{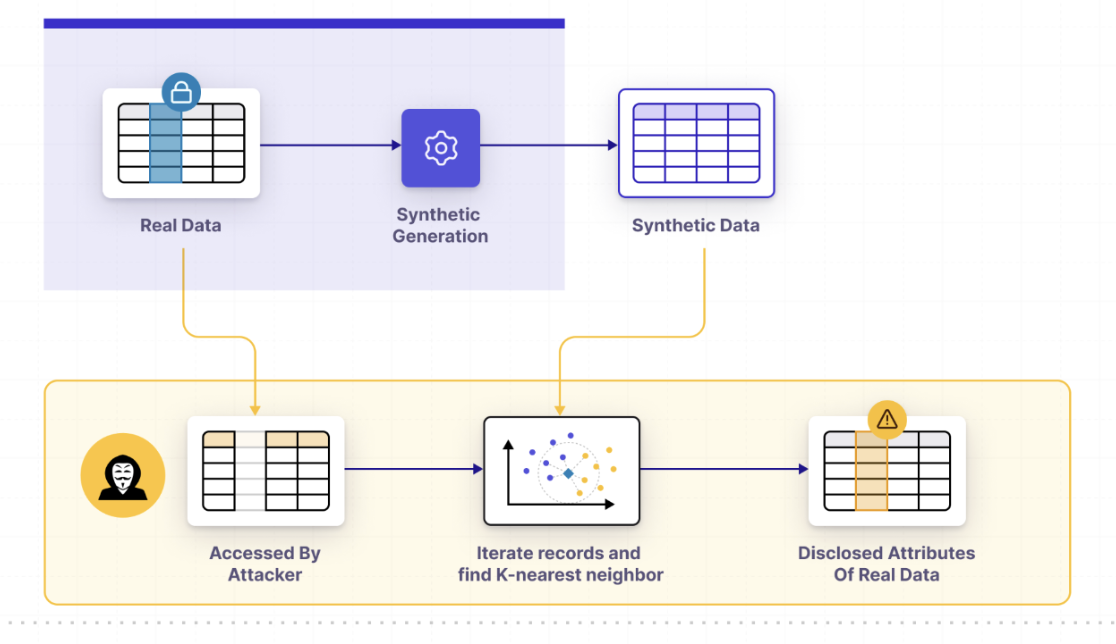}
\caption{KNN based AIA}
\end{figure}

The most common AIA is based on the k-nearest neighbors (KNN) algorithm \cite{59, 61, 65, 73, 74, 75, 76}. An overview of this AIA is shown in figure 2. In this AIA, it is assumed the attacker has access to the synthetic data as well as the quasi-identifiers for some portion of the real data. For each record in this attack dataset, the quasi-identifiers are used to find the k-nearest synthetic neighbors. 

From these k-nearest neighbors a mean record is computed. This is done by taking the mode for categorical fields and the mean for numeric. Then all the fields that are not quasi-identifiers are compared between the attack record and this mean nearest neighbor synthetic record. If the fields match, this is accumulated as a correct prediction for that column and if they don't match it's accumulated as an incorrect prediction. From these predictions you can get an AIA accuracy for each column. Sometimes column accuracy is weighted by the column's entropy as a way to take inherent predictability into account \cite{60, 65}. An overall score is often computed by taking the average of the column accuracies.

One variation of this common attack is, instead of KNN, a machine-learning model is trained (such as a random forest or gradient-boosted trees model) using the synthetic dataset S as a training set (where the quasi-identifiers Q are used as predictors and the sensitive variable t is used as the target) and then predict the value of t using this model \cite{54, 15}. Another variation is to train the distance score. For instance, Zhang et al \cite{77} propose a no-box attack against synthetic health records, using representation learning to train a similarity metric.

One challenge of using AIA as a privacy metric is there is no way to mathematically, unambiguously determine whether an attribute is an identifier, a quasi-identifier, or a non-identifying sensitive value. In fact, all values are potentially identifying, depending on their prevalence in the population and on auxiliary data that the attacker may have. Because of this, several research papers use the approach of randomly sampling which fields to use as quasi-identifiers \cite{65, 78, 79, 81}. Many attack records are processed and for each one some set number of random fields are used as the quasi-identifiers and the remaining fields are assumed to be sensitive (the fields we're trying to predict).

While AIAs are not as common as MIAs, new, unusual approaches are being published frequently \cite{55, 82, 83}. AIAs are a tangible, practical way to measure privacy risk.

\section{Privacy Enhancing Techniques}
It is often assumed that since synthetic data typically has no one-to-one correspondence with the original data, it will be sufficiently private. This assumption is not always valid, which has motivated the development of privacy metrics described in section 3. There are, however, a variety of mechanisms for strengthening the privacy of your synthetic data. Below are some of the best practices.

\subsection{Differential Privacy}

The most widely accepted approach to privacy is the use of differential privacy (DP) for training  synthetic data generation algorithms \cite{37,63,84,87}. Differential privacy is a mathematical definition of privacy. It provides the guarantee that an algorithm’s behavior hardly changes when a single record or entity joins or leaves the dataset. DP provides protection against adversarial attacks with high probability. A formal definition is as follows:

\textbf{Definition 1.} Differential Privacy: A randomized mechanism  \(\mathcal{M}\) with domain  \(\mathcal{D}\) (e.g. possible training datasets) and range  \(\mathcal{R}\) (e.g. all possible trained models) satisfies ($\epsilon$,$\delta$)-differential privacy if for any two adjacent datasets $d, d'\in D$ and for any subset of outputs  \(\mathcal{S}\)$\subseteq$ \(\mathcal{R}\) it holds that 
\[
\Pr[\mathcal{M}(d) \in S] \leq e^\epsilon \Pr[\mathcal{M}(d') \in S] + \delta
\]

Two datasets are considered adjacent if they differ by the existence of one training record. Hence the above definition is saying that the output of a model trained with DP (e.g. the synthetic data) should be roughly the same regardless of the existence of some specific user’s record in the training set.

In deep learning, DP is often achieved by using a differentially private stochastic gradient descent (DP-SGD) algorithm. DP-SGD achieves differential privacy by clipping gradients and adding noise during the optimization process. Other DP algorithms include Private Aggregation of Teacher Ensembles (PATE) \cite{85} and algorithms for categorical data, such as Private-PGM (Probabilistic Graphtical Models) and AIM (Adaptive and Iterative Mechanisms) \cite{110}. 

One of the challenges of using DP is maintaining reasonable  utility  of the data. This must be closely monitored when picking an $\epsilon$ to use. There is an interesting relationship between membershp inference attacks (MIAs) and DP. It has been shown that DP provides an upper bound on the effectiveness of MIAs and MIAs can be used to create a lower bound or effective $\epsilon$ \cite{88, 89, 90, 91, 92}.

\subsection{Pseudonymization}
Another privacy strengthening approach is to use pseudonymization of PII before training the synthetic generation model. Pseudonymization is a “data management and de-identification procedure by which personally identifiable information fields within a data record are replaced by one or more artificial identifiers, or pseudonyms” \cite{93}. Named-entity recognition (NER) is used to detect and replace PII present in  both structured and unstructured text fields \cite{94, 95}. Sometimes PII is masked instead of being replaced with pseudonyms. This is commonly referred to as "PII scrubbing.” PII scrubbing can mitigate membership inference but can lower utility much more than DP does \cite{9}.

\subsection{Privacy Filters}
Privacy filters can be used to eliminate synthetic records that are vulnerable to adversarial attacks. A similarity filter is a post-processing checkpoint that actively removes any synthetic data record that is overly similar to a training record, ensuring that no such record slips through the cracks, even in the case of accidental overfitting \cite{96, 97, 98}. An outlier filter is a post-processing checkpoint that actively removes any synthetic record that is an outlier with respect to the training data. Outliers revealed in the synthetic dataset can be exploited by membership inference attacks (MIAs), attribute inference attacks (AIAs), and a wide variety of other adversarial attacks \cite{96}.

While removing outliers from the synthetic data can improve privacy, this does not hold true for removing outliers from the original training data. Carlini et al \cite{97} describe a phenomenon called the  "Privacy Onion Effect" where removing the "layer" of outlier points that are most vulnerable to a privacy attack exposes a new layer of previously-safe points to the same attack.

Privacy filters can provide strong protection against data linkage, membership inference, and re-identification attacks with only a minimal loss in data accuracy or utility. However, they are not suitable for use when a DP guarantee is desired from the synthetic data as they utilize information from training records.

\subsection{Overfitting Prevention}
Many adversarial attacks exploit the fact that a generative model is overfit \cite{43, 55}. This means efforts to prevent overfitting will likely improve the privacy of synthetic data. Common practices to prevent overfitting include using batch-based optimization (e.g., stochastic gradient descent), regularization (e.g., dropout layers in deep neural networks), and early stopping. 

\subsection{Remove Duplication}
The success of MIAs has been shown to correlate with the presence of duplicated training data. Removing duplicates prior to training can improve the privacy of synthetic data \cite{99, 100, 101, 102, 103}.  Another approach is to preprocess the data to limit the number of records containing information about any one individual. The more records there are for an individual, the higher the likelihood of exposure to an adversarial attack. One must be careful, though, when deduplicating training data before applying differentially-private training as this has been shown to lead to adaptive privacy attacks \cite{109}.

\subsection{Data Augmentation}
Data augmentation involves creating new training data by modifying existing data, which can help reduce the reliance on sensitive data and minimize privacy risks. As the size of the training set increases, the success of MIAs decreases monotonically \cite{66, 105, 106}. This suggests that increasing the number of distinct samples in the training set can help to alleviate overfitting which then reduces the success of adversarial attacks.

\subsection{Use a Smaller Model}
Larger models are more prone to memorization. Hence an effective privacy technique is to reduce the size of the model you use, especially if it's an LLM. The performance of most types of MIAs increases as model size increases \cite{66, 100, 102, 104}. The exception is when Parameter-Efficient Fine-Tuning (PEFT) is used. This method enables efficient adaptation of large pretrained models to various downstream applications by only fine-tuning a small number of extra model parameters instead of all the model's parameters. Thus privacy is correlated with the number of parameters tuned as opposed to the overall model size. Note that using a smaller model could impact the utility of synthetic data.

\subsection{Enhance Architecture}
More sophisticated strategies have been further developed for different generative models. For example, privGAN \cite{107}, RoCGAN \cite{86} and PATE-GAN \cite{85} equip GANs with privacy protection through strategic changes to the model architecture. While these strategies exhibit enhanced effectiveness, they remain model-specific in nature, thereby limiting their broader applicability across various generative model paradigms.

\section{Conclusion}
Recent advancements in generative AI have made it possible to create synthetic data with accuracy that can be as effective as real-world data for training AI models, powering statistical insights, and fostering collaboration with sensitive datasets while offering strong privacy guarantees. Effectively measuring the privacy of synthetic data is an important step in the process. In this paper, we explored a wide variety of privacy metrics as well as different techniques for improving privacy. 

\bibliographystyle{unsrt}
\bibliography{privacy_metrics_arxiv}

\end{document}